# Human Identification using Selected Features from Finger Geometric Profiles

Asish Bera, and Debotosh Bhattacharjee, *Senior Member, IEEE*

*Abstract*— A finger biometric system at an unconstrained environment is presented in this paper. A technique for hand image normalization is implemented at the preprocessing stage that decomposes the main hand contour into finger-level shape representation. This normalization technique follows subtraction of transformed binary image from binary hand contour image to generate the left side of finger profiles (LSFP). Then, XOR is applied to LSFP image and hand contour image to produce the right side of finger profiles (RSFP). During feature extraction, initially, thirty geometric features are computed from every normalized finger. The rank-based forward-backward greedy algorithm is followed to select relevant features and to enhance classification accuracy. Two different subsets of features containing nine and twelve discriminative features per finger are selected for two separate experimentations those use the kNN and the Random Forest (RF) for classification on the Bosphorus hand database. The experiments with the selected features of four fingers except the thumb have obtained improved performances compared to features extracted from five fingers and also other existing methods evaluated on the Bosphorus database. The best identification accuracies of 96.56% and 95.92% using the RF classifier have been achieved for the right- and left-hand images of 638 subjects, respectively. An equal error rate of 0.078 is obtained for both types of the hand images.

## I. Introduction

BIOMETRIC systems are developed mainly for robust and secure human authentication. Increasing demand for higher security in diverse applications has directed researchers to explore several biometric traits to solve various challenging issues in the field of automated pattern recognition. Physiological traits, such as facial characteristics, palm print, fingerprint, hand geometric features; or behavioral traits like signature verification, and gait analysis are some well-known areas of research [1]. Hand geometry is regarded as one of the oldest biometric technologies [2]. Hand shape and its geometric characteristics can individualize a person from a large population [3]-[4]. The advantages of hand biometrics are mainly, lower cost of the sensor, lesser invasiveness, user-friendliness, and smaller template storage requirement [1]-[2]. Hand biometric systems are used in various commercial and government automated access control environment, such as automatic attendance maintenance [5]. In some forensic application, the quality of available fingerprint may be poor for recognition, in such circumstances the available hand geometry may be used for investigation.

Approaches based on hand biometrics, in unconstrained system, have attained research interests due to user flexibility and convenience. The most challenging problem in a contact-less environment is preprocessing, known as hand pose normalization [6]. A robust and viable pose-invariant normalization is essential to individualize the human identity [7]-[8]. Geometric attributes like lengths, widths, and palm area [9] are commonly measured from normalized fingers and/or hand for individualization. Alternatively, binary hand silhouette is also distinguishable using shape descriptors like Fourier descriptors (FDs) [7], scale invariant feature transform (SIFT) [10], wavelet coefficients [11]; and contour based matching such as independent component analysis (ICA)[2], [6], [12]; and shape-context (SC) [13]. Both of the geometric and shape-based features are equally essential and discriminative. Though, matching based on geometric features is rather easier than shape descriptors [14]. Importantly, key challenging issues during preprocessing are as follows:
i. Pose variation, inter-finger spacing, and hand accessories (e.g., bracelet, wristwatch) lead to artifact problems and create difficulties in finger tip and valley localization [6].
ii. Reliability on preprocessing depends on the constraints followed during image acquisition, such as the sensor, lighting conditions, and environmental factors [15].
iii. Exact finger tip-valley localization is an error-prone job, resulting in an inaccurate measurement of features. Thus, it increases intra-class discrepancy rather than inter-class variation and increases misclassification rate [2].

Due to inherent anatomic hand structure, pose variation is very sensitive. Thus, pose-invariant normalization necessitates few essential operations. Though deformable hand registration can be applied, it results in alignment error [16]. Another approach has been implemented at the finger-level, which is processing of a finger locally. It has been studied that preprocessing of the thumb, as well as finding out its extreme points are somehow complex [13]. Particularly, the angular deviation of the thumb and in-between space of the thumb and index finger are inconsistent due to pose flexibility and anatomic foundation. In literature, some works have been experimented

A. Bera is with the Department of Computer Science and Engineering, Haldia Institute of Technology, Haldia-721657, India
(e-mail: asish.bera@gmail.com)
D. Bhattacharjee is with the Department of Computer Science and Engineering, Jadavpur University, Kolkata-700032, India
(e-mail: debotosh@ieee.org)

TABLE I
IMPORTANT CHARACTERISTICS OF APPROACHES BASED ON POSE-INVARIANT HAND BIOMETRICS USING RIGHT HAND (RH), AND/OR LEFT HAND (LH) IMAGES.

| Sl. | Methodology | Experimentation | Remarks | Ref. |
|---|---|---|---|---|
| 1 | *Shape-based (four fingers):* Cumulative angular function (CAF) based Fourier descriptors (FDs) from finger contour and finger area are fused at score-level. | *DB:* Bosphorus (638, LH). *Fusion:* score level | No identification accuracy is reported. Modifications over traditional segmentation and feature extraction methods are applied. | [7] |
| 2 | *Shape-based (four fingers):* Coherent distance shape contexts (CDSC) and based on shape contexts (SC) and inner-distance shape contexts (IDSC). | *DB:* CASIIM (200, LH). *Fusion:* score level. | Computing time of the method is a little bit long for real application. | [13] |
| 3 | *Shape-based (five fingers):* Various shape and appearance based experiments are conducted using ICA features of left and/or right hands. | *DB:* Bosphorus (800, both hands) *Fusion:* feature & score level. | Modified normalization of the method presented in [6] is followed; feature dimension is very high up to 600. | [2] |
| 4 | *Shape-based (five fingers):* ICA features using complex angular radial transform (ART), denoted as shape_ART. | *DB:* Bosphorus (756, both hands) | Complex due to ART and feature dimension is high. | [3] |
| 5 | *Shape-based (five fingers):* ICA on normalized binary hand to extract and summarize prototypical shape information. | *DB:* Bosphorus (458,RH) | Statistical ICA feature dimension is high, at least 300 for prototyping hand shape. | [6] |
| 6 | *Shape-based (five fingers):* SIFT features are invariant to scale, illumination, rotation, noise, and distortion. SIFT matching based on the key points is used. | *DB:* IITD (234, LH), Bosphorus (642, RH) | Complexity depends on the Gaussian linear transformation and searching of invariant points at different Gaussian scale space. SIFT feature space is high. | [10] |
| 7 | *Geometric and Shape-based (five fingers):* 1-D wavelet decomposition at level-5 using Daubechies-1 wavelet filter is applied. From each of the distance map and orientation map, first 50 decomposed wavelet coefficients along with 21 geometric features are computed. | *DB:* JUET (50, RH) and IITD (240, RH). *Fusion:* two levels of matching score: Level-1, & Level-2. | No identification accuracy is reported; two levels of score-level fusion improve EERs; population is low and feature space is high. | [11] |
| 8 | *Geometric and Shape-based (five fingers):* Normalization is similar to [6]; except it describes iterative closest point algorithm for modified Hausdorff distance measure. | *DB:* Bosphorus (500, RH) *Fusion:* score level. | Each ICA feature vector contains higher number of data points as 30720, weighted score fusion improves accuracy. | [12] |
| 9 | *Geometric and Shape-based (five fingers):* Genetic Algorithm (GA) with a fitness function is applied for feature selection. MI is applied to find out the correlation between a pair of features and to reduce redundancy among features. | *DB:* GPDS(144, RH), IITD (137, unspecified), and CASIA (100, unspecified) | Selects about 50 features from 403 features based on 100 executions of GA. Dataset sizes are limited within 150 subjects. | [14] |

with four fingers, ignoring the thumb while other works have considered all the five fingers. Some state-of-the-art methods are summarized in Table I. Moreover, research interests are concentrating on the 3D hand geometry, fusion of 2D and 3D hand features [5], [8]; 3D palmprint [17], thermal hand images [15], [18]-[19]; and synthetic hand images [20]-[21].

Extraction of various kinds of features is not sufficient to achieve good performance. Feature selection algorithm (e.g., filter, wrapper) plays an important role in pattern recognition to choose discriminating features from a high dimensional feature space to improve classification accuracy and thereby reduce computational time [22]-[27]. Selection of relevant features determines a good combination of attributes that can reduce classification error by maximizing relevance and minimizing redundancy in the sample feature space [25]. An Information Theory metric, such as the mutual information (MI), is used to select relevant features for individualization in the context of hand biometrics [14], and gait recognition [28].

In this work, mainly, an approach based on four fingers excluding the thumb in a contact-free scenario has been emphasized[†1]. A new normalization method based on the gradient magnitude variation of the hand image and Boolean operations has been implemented to extract the finger shape profiles. Geometric features are computed from normalized fingers as those are simpler to compute and avoid difficulties related to shape-based features. Locating tip-valley point is not essential to compute the features for a method that uses this normalization. Next, discriminative and reliable feature selection is necessary for hand biometric system with a larger population.

Here, the advantages of the sequential forward and backward selection methods [29] are considered in an adaptive forward-backward feature selection algorithm to improve the accuracy. To deal with feature overfitting issue [26], optimal subsets of the relevant features are validated accordingly. The main contributions of this work can be summarized as:

i. *Pose-invariant hand image normalization by the arithmetic and logical operations.* This normalization uses subtraction and Boolean operations on binary hand images to extract shape profiles and every finger can be separated from each other by this normalization in a contact-less environment.

ii. *Selection of relevant features to reduce computational cost.* A rank-based adaptive forward-backward feature selection method is followed to find out the relevant features from a larger feature space.

iii. *Removal of redundant features to improve classification accuracy.* Elimination of redundant features enhances the correctness for individualization.

iv. *Noting the performance of the present technique through experiments conducted once including thumb i.e., five fingers and another excluding thumb i.e., four fingers only.* The effectiveness of each finger in individualization is evaluated based on the optimized feature subsets.

Experiments are conducted on the *Bosphorus* database with the right- and left-hand images of 638 subjects to assess the performance by the kNN and RF[30]-[31] classifiers.

The remainder of this paper is organized as follows: the preprocessing method is described in Section II. Feature description and relevant feature selection are presented in Sec-

---

[†1]*Note: In this paper, four fingers represent all the fingers except the thumb finger.*

tion III. Experimental illustrations with results are discussed in Section IV, and the conclusion is drawn in Section V.

## II. Hand Image Normalization Methodology

This pose-invariant normalization decomposes the hand silhouette to segment each finger using simple arithmetic and Boolean operations. Every finger contour is divided into two parts, containing the left-side and right-side of the finger shape profiles (FPs). Finally, these two contour segments are merged to represent a finger which is separated from others. The normalization method is divided into three stages, shown in Fig.1.

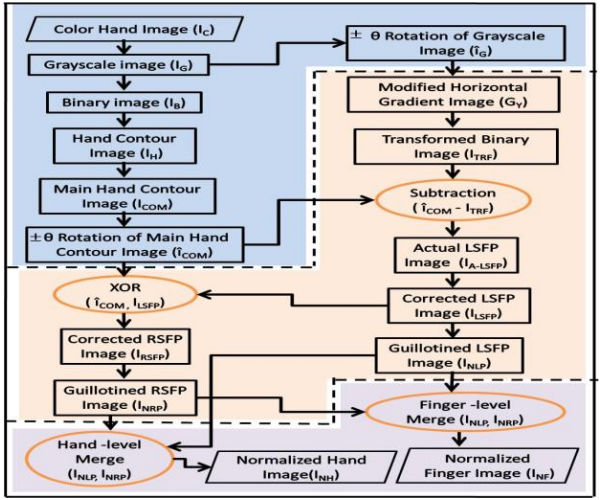

Fig.1. Hand image normalization methodology.

*Stage 1: Basic image preprocessing*

Firstly, a color hand image $I_C$ is converted into its grayscale equivalent image $I_G$. Noise can be introduced during image acquisition which may cause variations in pixel intensities abruptly. Thus, noise has been removed by the nonlinear spatial filter, namely the median filter [14]. Next, $I_G$ is converted into a binary image by the global thresholding (*th*) which is determined using the Otsu's method [32] as

$$I_B = T_B(I_G, th) \qquad (1)$$

where the threshold is denoted by *th* and its value lies between {0,1}. Morphological operator is applied to fill smaller holes and foreground blobs in $I_B$. It is necessary to smooth the hand silhouette, to recover missing contour pixels, and to remove spurious pixels. Hand contour image $I_H$ is determined from $I_B$ using the Canny's edge detection algorithm [33]. Canny's edge detector uses the Gaussian filter to remove noise. Mainly, it considers the gradient magnitude image and double threshold values to determine the optimal solution. It follows an edge thinning method, namely the non-maximum suppression on the gradient magnitude image to remove the spurious edges. Then, double threshold values are used to select only the potentially strong edges. In this work, FPs segmentation is performed based on the modified gradient magnitude, described at S*tage 2*. Image $I_H$ may contain multiple disjoint components due to several reasons, such as dirt artifact on hand surface, clothing, ornament, and intensity variation. The largest connected foreground component $I_{COM}$ is selected from $I_H$ based on the maximum area, known as area-based filtering [6]. Similarly, the smaller and unnecessary components are removed which contain a very lesser number of pixels, mainly found at the wrist region and sleeves due to clothing artifacts. Next, the region of interest (ROI) is determined by selecting the minimum bounding rectangle (MBR) of the main component $I_{COM}$. During preprocessing in a contact-free system, a normal orientation i.e., rotation of ROI is essential to avoid angular dissimilarity. Posture uniformity by rotation is achieved using an ellipse fitted over the ROI so that the major axis of the ellipse passes through the centroid of ROI. The centroid is defined as

$$(x_c, y_c) = (m_{1,0}/m_{0,0}, m_{0,1}/m_{0,0}) \qquad (2)$$

where, $m_{i,j}$ is the moment of an image. Now, $I_{COM}$ is rotated with respect to its centroid.

$$\hat{I}_{COM} = I_{COM} \times [T_R]_\theta \qquad (3)$$

where, $[T_R]_\theta$ represents 2D transformation matrix for rotation. The angle of rotation is given as

$$\theta = 0.5 \tan^{-1}\left(\frac{2m_{1,1}}{m_{2,0} - m_{0,2}}\right) \qquad (4)$$

The major-axis should be coincident with the coordinate Y-axis, as followed in [11], [34]. The inertia matrix can be envisaged as the fitted ellipse over the ROI and the larger eigenvalue corresponds to the major-axis. The direction of rotation is followed towards the larger eigenvector. The rotation method is akin to as pictorially shown in [11]. Similarly, before further processing, $I_G$ is also rotated with the same angle ($\theta$) as

$$\hat{I}_G = I_G \times [T_R]_\theta \qquad (5)$$

*Stage 2: Image transformation and shape profiles extraction*

It is the most critical stage of normalization algorithm for FPs segmentation. It is divided into three steps, namely, the image transformation, FPs extraction, and wrist removal.

*A. Image transformation*

Local intensity-profile variation between every pair of neighborhood pixels of $\hat{I}_G$ is computed, and denoted by image $G_Y$. A modified operation akin to computing the horizontal gradient image based on the first order derivative is followed. This operation generates image $G_Y$ and defined as

$$G_Y \cong \hat{I}_G \otimes \Delta_F = [\hat{I}_G(i,j) - \hat{I}_G(i, j+1)]/Þ \qquad (6)$$

where $i$ represents the row-index, $j$ is the column-index, and $Þ$ is a constant ranging from one to the maximum gray level intensity value i.e., $1 \leq Þ \leq 255$. Eq.6 represents the horizontal gradient magnitude when $Þ=1$. All the pixels of $G_Y$ turned into black i.e., intensities become zero, when $Þ=255$. This operation has been followed according to the direction of the Cartesian X-axis which is represented as a column of a 2D image. Intensity-profile variation of $G_Y$ is richer at lower values of $Þ$. On the contrary, a higher value of $Þ$ captures only higher intensity variations and certain essential pixels from the ROI boundary having comparatively lower intensities are left out. As a result, at the next step (Eq.7), FPs are segmented into multiple disconnected components due to the omission of edge pixels at higher values of $Þ$. Consequently, the true FPs cannot be retrieved properly after subtraction followed in Eq. 13. Local intensity variation in $\hat{I}_G$ is controlled by $Þ$ and for $Þ > 2$ or even higher up to $Þ = 8$, $G_Y$ represents intensity variations only at the right side of FPs. Hence, to extract good shape profiles, the value of $Þ$ should be adjusted according to intensity profiles of the raw image. It is noted that $Þ=1$ or $Þ=2$ produces similar $G_Y$ images. A higher value of $Þ$ causes elimination of the extreme pixels which is very sensitive to pose alternation and it causes to remove the end pixels at the tip-valley region. It results in intra-class variations during feature extraction. After several experiments the minimum value, $Þ=2$ is chosen.

The simplest convolution operator to compute $G_Y$ is $\Delta_F = \{1, -1\} \approx \{1,0,-1\}$. Mainly, $G_Y$ considers high- to low-intensity variation between every two nearest pixels horizontally and this variation occurs mostly at the right-side of FPs in $\hat{i}_G$ (Eq.6). Alternatively, changing the convolution operator $\Delta_F = \{-1,1\} \approx \{-1,0,1\}$ results in reverse outcome, i.e., low- to high-intensity variation occurs at the left side of FPs. The transition from high- to low-intensity profile, i.e., foreground pixel (1-valued) of finger shape to background pixel (0-valued) occurs at the right-side of a finger. Similarly, low- to high-intensity variation occurs during the transition from background to foreground pixel. Every $G_Y$ image is represented as an ordered pair of pixels and corresponding intensities, i.e., $G_Y = \{(i,j), f(i,j)) / i=1,2,..,M; j=1,2,..,N; f(i,j)= 1,2,...,255\}$; where $M \times N$ is the dimension of $G_Y$. Image $\hat{i}_G$ is transformed linearly to incorporate local changes of pixel intensity. Now, $G_Y$ is converted into a binary image based on the intensity of every pixel and named as the transformed image $I_{TRF}$.

$$I_{TRF}(i,j) = \begin{cases} 1 & \text{if } G_Y(i,j) \geq 1 \\ 0 & \text{otherwise} \end{cases} \quad (7)$$

Intensity deviations of $I_{TRF}$ from high to low are represented by white pixels, and remaining pixels are converted into the black. The white pixels of $I_{TRF}$ are located mainly at the right-side of fingers. Some of those white pixels constitute the true edge to represent the FPs. While the remaining other non-zero pixels are denoted as $\psi$. Due to noise and abrupt intensity variation, $\psi$ may contain white pixels that span over any spatial location of $I_{TRF}$ (Fig.3.c). Thus, $I_{TRF}$ can be defined as $I_{TRF} = \{I_{RSFP} \text{ OR } \psi\}$, where $I_{RSFP}$ represents the right side of all finger profiles (RSPF) and $\psi$ denotes the remaining non-zero pixels.

### B. Finger shape profiles extraction

The foreground pixels of the rotated hand contour $\hat{i}_{COM}$ (Eq.3) can be logically decomposed and represented (Fig.2.a) into two binary images, namely the $I_{LSFP}$ and $I_{RSFP}$. The non-zero pixels that represent the left-side of FPs (LSFP) of all fingers are denoted as $I_{LSFP}$ (Fig.3.g). Image $I_{RSFP}$ (Fig.3.h) represents the right-side of all FPs (RSFP). Altogether, $I_{LSFP}$ contains five disconnected contour components, one per finger that represents LSFP. Similarly, $I_{RSFP}$ represents the other side of those five disjoined contour segments. The inclusion of corresponding one-to-one LSFP and RSFP represents the respective FP which is segmented from other fingers (Fig.3.j). Thus, based on the FPs, $\hat{i}_{COM}$ can be represented as

$$\hat{i}_{COM} = \{I_{LSFP} \text{ OR } I_{RSFP} \mid I_{LSFP} \text{ AND } I_{RSFP} = 0\} \quad (8)$$

It implies that all the contour pixels are spatially and mutually exclusively exist in either $I_{LSFP}$ or $I_{RSFP}$ so that the logical OR operation between the two images can represent the image $\hat{i}_{COM}$. Image $I_{TRF}$ has been defined earlier as

$$I_{TRF} = \{I_{RSFP} \text{ OR } \psi\} \quad (9)$$

Now, subtraction is followed between $\hat{i}_{COM}$ and $I_{TRF}$ images.

$$\begin{aligned} I_{SUB} &= \hat{i}_{COM} - I_{TRF} \\ &= \{I_{LSFP} \text{ OR } I_{RSFP} \mid I_{LSFP} \text{ AND } I_{RSFP} = 0\} - \{I_{RSFP} \text{ OR } \psi\} \\ &= I_{LSFP} - \psi \\ &= I_{LSFP} \end{aligned} \quad (10)$$

After subtraction, the spatial locations of $\psi$ with negative or zero intensity are turned into black. Hence, it results in only the LSFP. The LSFP image can be defined as

$$I_{LSFP}(i,j) = \begin{cases} 0 & \text{if } I_{SUB}(i,j) \leq 0 \\ 1 & \text{otherwise} \end{cases} \quad (11)$$

Subtraction (Eq.10) generates $I_{LSFP}$ which defines only LSFP of a hand contour including all fingers. However, this actual LSFP, denoted as $I_{A-LSFP}$, may contain multiple smaller components mainly at the wrist region (Fig.3.f). Subtraction may also cause a discontinuity in LSFP. Thus, few missing pixels which cause a discontinuity in LSFP are recovered using morphological closing with $3 \times 3$ square structure element. Bridging between the same disconnected FP is performed. Finally, thinning operation is applied to represent FPs as thin as possible. Then, five FP components, one per finger are sorted out according to their positions i.e., from right to left direction of a right-hand. It is considered as the corrected LSFP[†2], denoted by $I_{LSFP}$. Finally, XOR is applied between $\hat{i}_{COM}$ and $I_{LSFP}$.

$$\begin{aligned} I_{XOR} &= \{\hat{i}_{COM} \text{ XOR } I_{LSFP}\} \\ &= \{\{I_{LSFP} \text{ OR } I_{RSFP} \mid I_{LSFP} \text{ AND } I_{RSFP} = 0\} \text{ XOR } I_{LSFP}\} \\ &= I_{RSFP} \end{aligned} \quad (12)$$

Logical XOR considers the contour pixels of either $\hat{i}_{COM}$ or $I_{LSFP}$, but not both. Hence, $I_{XOR}$ represents the corrected $I_{RSFP}$.

### C. Wrist region removal

The lower portion of little finger LSFP and thumb RSFP are associated with palm and wrist. The size of a hand with wrist area varies according to clothing occlusion, hand gadget, and pose variation from intra-person to inter-person. Thus, this region must be eliminated. A fixed vertical distance of 100 pixels from the centroid of every hand has been considered for removing wrist in [7]. Though, wrist smoothing method based on fixed height is not robust in all cases, particularly, where alignment of the thumb is huge. Hence, it may cause an error in feature computation. Here, wrist irregularity along with a lower portion of palm is removed using a reference line which is automatically determinable for every image and varies accordingly (if necessary). This method is shown in Fig.2. The end point of thumb LSFP is considered as a pivot point $P_1$. From $P_1$, vertically downwards scanning is done on the same column of the thumb RP. The first intersecting non-zero pixel $P_2$ on the thumb RP is determined. Now, from $P_2$ searching is done horizontally for the non-zero pixel from right to left direction, i.e., from thumb towards little finger to find another intersecting point $P_3$ on the little finger LP. Pixels $P_2$ and $P_3$ are marked as end points of the reference line. The lower portions of LSFP and RSFP which exist below the reference line are eliminated. After wrist removal, fingers are normalized into a standard orientation. Now, the normalized $I_{LSFP}$ and $I_{RSFP}$ are denoted by $I_{NLP}$ and $I_{NRP}$, respectively.

### Stage 3: Finger shape isolation

Images $I_{NLP}$ and $I_{NRP}$ can be merged in two ways.

---

[†2]Note: In ideal case the actual LSFP ($I_{A-LSFP}$) and the corrected LSFP ($I_{LSFP}$) represent the same image. In that case, the morphological operations are not essential. Hence, both images are denoted by $I_{LSFP}$.

(i) *Global or hand level*: $I_{NLP}$ and $I_{NRP}$ are merged to represent the normalized hand shape $I_{NH}$.

(ii) *Local or finger level*: the merging of one-to-one LP and RP components represent every single normalized finger $I_{NF}$. Images $I_{NLP}$ and $I_{NRP}$ can be represented at finger-level as

$$I_{NLP} = \{LP_{Thumb}, LP_{Index}, LP_{Middle}, LP_{Ring}, LP_{Little}\} = \{LP_i\}_{i=1}^{5} \quad (13)$$

$$I_{NRP} = \{RP_{Thumb}, RP_{Index}, RP_{Middle}, RP_{Ring}, RP_{Little}\} = \{RP_i\}_{i=1}^{5} \quad (14)$$

where, $LP_i$ and $RP_i$ indicate the left- and right- profiles of each finger, respectively. Similarly, finger-level representation can be given as

$$I_{NF} = \{LP_i \text{ OR } RP_i \,/\, LP_i \text{ AND } RP_i = 0\}_{i=1}^{5}$$

$$= \{F_{Thumb}, F_{Index}, F_{Middle}, F_{Ring}, F_{Little}\} = \{F_i\}_{i=1}^{5} \quad (15)$$

where, $F_i$ represents every normalized finger, denoted as $I_{NF}$ (Fig.3.j).The number of pixels in either $LP_i$ or $RP_i$ shape profile is asymmetric and inconsistent. However, it is not an important factor because this work is based on the individual finger. Features are computed at the finger-level i.e., local to every $I_{NF}$. Now, every $I_{NF}$ is fixed at 280×160 pixels.

---

**Algorithm 1**: *finger profiles (LSFP and RSFP) retrieval*

---

**Input:** *Main contour image $\hat{I}_{COM}$ and grayscale image $\hat{I}_G$*
**Output:** *Corrected shape-profile images $I_{LSFP}$ and $I_{RSFP}$*

1. **for** each pixel of image $\hat{I}_G(i,j)$ compute the modified the horizontal gradient image
   $G_Y \leftarrow (\hat{I}_G(i,j) - \hat{I}_G(i,j+1))/P$
   **end for**
2. transform $G_Y$ into the binary image $I_{TRF}$ using Eq.7
3. perform binary subtraction to produce actual LSFP
   $I_{SUB} \leftarrow \hat{I}_{COM} - I_{TRF}$
4. **for** each pixel of image $I_{SUB}(i,j)$
   **if** $I_{SUB}(i,j) \leq 0$ **then**
   $I_{LSFP}(i,j) \leftarrow 0$
   **else**
   $I_{LSFP}(i,j) \leftarrow 1$
   **end if**
   **end for**
5. apply morphological operation on the actual LSFP and arrange them according to positional sequences of fingers to obtain the corrected $I_{LSFP}$ image
6. apply XOR to compute the corrected RSFP image
   $I_{XOR} \leftarrow \hat{I}_{COM} \text{ XOR } I_{LSFP}$
7. **end**

---

The complexity of this algorithm depends on the operations for normalization over the input image. If the dimension of the input image is $M \times N$ pixels, then the complexity is $O(M \times N)$.

## III. FEATURE DEFINITION AND SELECTION

During feature computation, altogether thirty geometric features are computed from every $I_{NF}$. Then, a feature selection method has been followed to choose optimal subsets of relevant features. Two optimal subsets are defined and the second subset is a subset of the first subset. The second subset contains the least number of features with a minimum reduction of accuracy. However, further minimizing the number of features causes significant performance degradation.

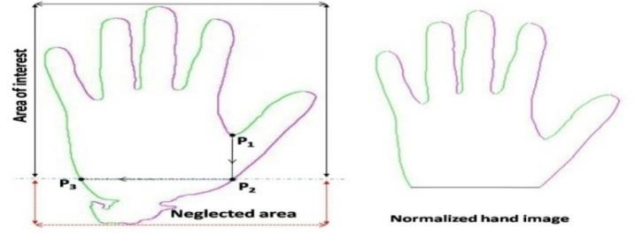

Fig.2. Wrist irregularity removal for smoothness

### A. Feature Definition

Three different sets of features per finger are computed, namely set $A_1$, $A_2$, and $A_3$. The features are defined as follows:

**Set $A_1$:** Ten important geometric descriptors per finger are calculated. Some features such as area, perimeter, solidity, and others are also defined in [14].

(i) **area:** the number of boundary pixels of a finger, denoted as *FINGERAREA*. (ii) **perimeter:** the distance around the finger boundary, calculated as the distance between each pair of adjacent contour pixels of a finger. (iii) **major-axis length** and (iv) **minor-axis length**: the major-axis and minor-axis lengths of an ellipse which is fitted on each finger. These axes lengths are lesser susceptible to noise, and invariant to affine transformations like translation and rotation to a large extent. (v) **length** and (vi) **width** of the *MBR*. (vii) **filled area:** the number of pixels corresponds to a finger with all holes filled in providing the same *MBR*. (viii) **equivalent diameter:** diameter of a circle with the same area, computed as $(4 \times AREA/\pi)^{1/2}$ (ix) **solidity:** the proportion of the contour pixels in the convex hull, calculated as *FINGERAREA/CONVEXAREA* and (x) **extent:** the ratio of pixels in the finger to pixels in the entire *MBR* given as *FINGERAREA/MBRAREA*.

**Set $A_2$:** Every $I_{NF}$ is regarded as a closed contour by joining two end valley points. The centroid local to the finger is determined. Now, assuming the closed contour as a curve, boundary pixels $(x_i, y_i)$ are labeled from the left-side valley as the starting point. According to the arc length parameterization within $\{0,1\}$, equal-distance points are located. Distances from the centroid $(x_c, y_c)$ of the finger to ten different contour points which are located equal distance apart are computed. The centroidal distance $D$ from $i^{th}$ point $(x_i, y_i)$ is defined as

$$D_i = [(x_i - x_c)^2 + (y_i - y_c)^2]^{1/2} \quad (16)$$

The $D_i$ distances are translation-invariant feature [35].

**Set $A_3$:** Finger widths at ten different equidistant positions along the length of every finger are calculated and included in another feature set.

### B. Rank-based adaptive forward-backward feature selection

A feature can be either relevant or irrelevant [22], [26]. Again, the relevance has been defined as strongly relevant or weakly relevant. A feature is relevant when it dominates over other features regarding the accuracy. Feature relevance and redundancy are defined here in the context of forward and backward incremental search method. Let, the input set *A* is a

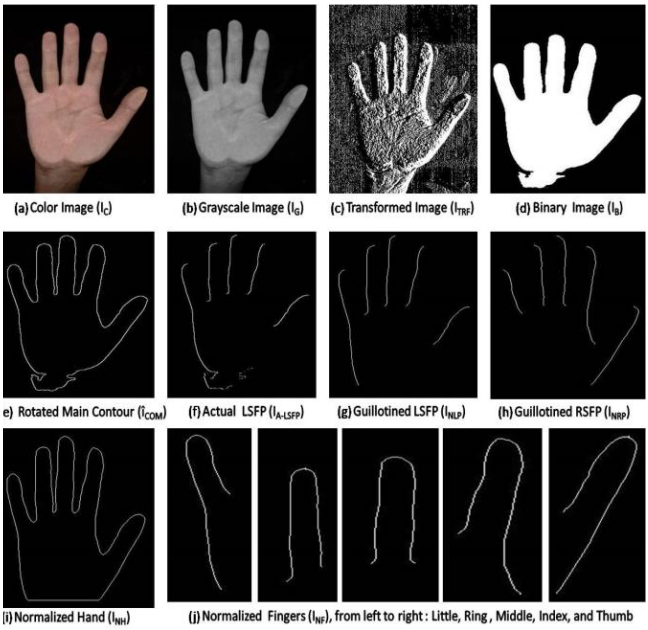
Fig.3. Outcomes subsequent steps of hand image normalization

$w$-dimensional feature vector, and each feature is denoted by $f_i$. Each vector is associated with a class label $h_k$ in the set of all classes $H$. The subset of selected features is $B$ and its classification accuracy is denoted as $\mathfrak{S}(B)$.

**Definition 1**: A feature $f_i$ is called relevant to the class $h_k$ in the sample space, if $\mathfrak{S}(B \cup f_i) > \mathfrak{S}(B)$ or $\mathfrak{S}(B \backslash f_i) < \mathfrak{S}(B)$ otherwise, $f_i$ is irrelevant.

A relevant feature improves classification accuracy with the selected subset of attributes at any stage, and it is correlated with the target class. Relevance is determined by including $f_i$ into subset $B$ in forward search; and in backward search, $f_i$ is taken away from $B$ i.e., $B-\{f_i\}=B \backslash f_i$. Consequently, irrelevant features are removed. Sometimes, an irrelevant feature degrades the performance.

**Definition 2:** A feature $f_i$ is called redundant to the class $h_k$ in the sample space, if $\mathfrak{S}(B \cup f_i) = \mathfrak{S}(B)$ or $\mathfrak{S}(B \backslash f_i) \geq \mathfrak{S}(B)$.

During the forward step, redundancy is determined by the additive process and $f_i$ is insignificant to subset $B$ regarding the accuracy. It does not add or reduce any gain to the target class.

In backward step, redundancy is found through feature removal process. Deletion of a redundant feature may improve the accuracy. Thus, during backward elimination not only redundant features are removed, but also further improvement of classification accuracy is possible.

Here, the selected feature subspace is global in the sense that a chosen feature $f_i$ is relevant to all the classes in $H$. Subset $B$ should contain all the relevant features, from which removal of any feature results in degradation of classification accuracy. Our objective is to find out the best possible combination of relevant features from a given feature set to achieve the maximum accuracy and minimum error. Thus, an optimal feature subset must be formulated that improves the accuracy.

The forward and backward greedy algorithms are two popular filter-based methods for selecting relevant features from a given high-dimensional feature space that may also contain some noisy data present. However, these algorithms have explicit shortcomings. There is no scope to correct errors made at initial steps in forward selection. Backward elimination leads to high computational cost, as it starts with all attributes. These two methods are combined to overcome their limitations and enhance the effectiveness, known as an adaptive forward-backward greedy algorithm, denoted as *FoBa* in [29]. In Algorithm-2, firstly, the features are sorted according to the relevance. Based on the order of relevance, a rank is assigned to every feature. Next, forward selection is performed sequentially according to the rank. Then, greedy backward elimination is applied that ensures an improvement over forward selection in terms of accuracy. The rank criterion is imposed on testing the features from higher- to lower-order of relevance while evaluating the subsets of selected features[†3]. Rank-based selection can also reduce the scope of redundancy of weakly relevant features by considering highly relevant features at the early stages. In this work, Algorithm-2 is named as the *Rank-based forward-backward* (*RFoBa*) feature selection algorithm.

---

**Algorithm 2**: *RFoBa feature selection algorithm*

---

*Input:* Feature set: $A$
*Output:* Optimal feature subset: $B_{OPT}$

1. *features are arranged according to the rank, $A \leftarrow \{f_1, f_2, .., f_w\}$; initialize, the number of features during forward selection, $k \leftarrow 0$, the number of optimal features $p \leftarrow 0$, and the selected feature subset, $B_{SEL} \leftarrow \emptyset$.*
2. *select the first feature $f_1$ which has the maximum relevance $B_{SEL} \leftarrow f_1$ and $A \leftarrow A \backslash \{f_1\}$;*
   $k \leftarrow 1$
3. **Repeat** *until no more feature is added in $B_{SEL}$*
   **for** $i \leftarrow 2, 3, ...., w$
     **if** $f_i \notin B_{SEL}$
       **if** $\mathfrak{S}(B_{SEL} \cup f_i) > \mathfrak{S}(B_{SEL})$
         $B_{SEL} \leftarrow B_{SEL} \cup \{f_i\}$ and $A \leftarrow A \backslash \{f_i\}$
         $i \leftarrow i+1$
         $k \leftarrow k+1$
       **end if**
     **end if**
   **end for**
4. *after a forward search, $k$ features are selected in $B_{SEL} \leftarrow \{\hat{f}_i\}^k_{i=1}$ which would be used for backward elimination in the next steps.*
   $B_{OPT} \leftarrow B_{SEL}$; $p \leftarrow k$;
5. **for** $i \leftarrow 1, 2, ...., k$
     **if** $\mathfrak{S}(B_{OPT} \backslash \hat{f}_i) \geq \mathfrak{S}(B_{OPT})$
       $B_{OPT} \leftarrow B_{OPT} \backslash \{\hat{f}_i\}$
       $i \leftarrow i+1$
       $p \leftarrow p-1$
     **end if**
   **end for**
6. *return $B_{OPT}$ with $p$ optimal features.*
7. **end**

---

Forward selection starts with an empty subset $B^{(0)}_{SEL}$. Firstly, classification accuracy $\mathfrak{S}(A)$ is computed with all the $w$ fea-

---

[†3]*Note: after forward selection $\hat{f}_i$ is used to denote the new position or index of $f_i$ in $B_{OPT}$.*

TABLE II
ACCURACY (%) OF EACH FEATURE OF RIGHT HAND (RH) AND LEFT HAND
(LH) USING kNN BASED ON THE AVERAGE SAMPLE OF 100 SUBJECTS

| Feature Set $A_1$ | RH | LH | Set $A_2$ | RH | LH | Set $A_3$ | RH | LH |
|---|---|---|---|---|---|---|---|---|
| $a_1$: Area | 62 | 47 | $b_1$ | 43 | 35 | $c_1$ | 29 | 31 |
| $a_2$: MBR length | 25 | 23 | $b_2$ | 36 | 31 | $c_2$ | 32 | 29 |
| $a_3$: MBR width | 67 | 66 | $b_3$ | 72 | 41 | $c_3$ | 33 | 41 |
| $a_4$: Major-axis len. | 74 | 61 | $b_4$ | 30 | 37 | $c_4$ | 33 | 32 |
| $a_5$: Minor-axis len. | 79 | 65 | $b_5$ | 35 | 31 | $c_5$ | 41 | 37 |
| $a_6$: Filled area | 62 | 47 | $b_6$ | 44 | 39 | $c_6$ | 37 | 37 |
| $a_7$: Equiv. diameter | 61 | 48 | $b_7$ | 31 | 28 | $c_7$ | 41 | 36 |
| $a_8$: Perimeter | 77 | 70 | $b_8$ | 44 | 41 | $c_8$ | 42 | 45 |
| $a_9$: Solidity | 41 | 35 | $b_9$ | 39 | 34 | $c_9$ | 39 | 38 |
| $a_{10}$: Extent | 24 | 22 | $b_{10}$ | 25 | 24 | $c_{10}$ | 30 | 31 |

tures. The accuracy of each feature is calculated independently (Table II), and a rank is assigned accordingly. The most relevant feature carries the highest accuracy, and its rank is

$$r_1 = \arg\max_{f_i \in A} \mathcal{S}(f_i), \text{ where, } i=1,2,...,w. \quad (17)$$

Similarly, all features are arranged in descending order of relevance, i.e., according to the rank $A=\{f_1,f_2,...,f_w\}$. Say, the $j^{th}$ feature is the most relevant one and ranked as 1, denoted as $f_j^{(1)}$. For simplicity $f_j^{(1)}$ is labeled as $f_1$. Now, $f_1$ is included in $B^{(1)}_{SEL}$ as the first element. Next, $f_2$ is paired with $f_1$ and the accuracy is evaluated. If the accuracy of combined $f_1$ and $f_2$ is better than that of the only $f_1$, then $f_2$ is included in $B^{(2)}_{SEL}$, otherwise not. This incremental process is repeated for all the remaining features until no more features are included, $f_e \in A$ where, $e=3,...,w$. Let, $k$ features are selected as

$$B^{(k)}_{SEL}=\{f_1,f_2,..,f_k\} \text{ and } B^{(k+1)}_{SEL} \leftarrow B^{(k)}_{SEL} \cup \{f_{k+1}\} \quad (18)$$

Where $1<k<w$. If $\mathcal{S}(B^{(k+1)}_{SEL})>\mathcal{S}(B^{(k)}_{SEL})$ then $f_{k+1}$ is accepted; otherwise, $f_{k+1}$ is rejected as irrelevant. After the first iteration, the rejected features are considered again one at a time and tested with a target of further probable gain in accuracy. In this case, several different combinations of features have been tested. During the first iteration, if $f_k$ is rejected, it is again included in $B_{SEL}$ with a different sequence of selected features and tested. Thus, step 3 has been repeated until no more features can be included in $B_{SEL}$. After forward selection, $B^{(k)}_{SEL}$ contains $k$ relevant features and remaining irrelevant features are discarded. Now, $\mathcal{S}(B^{(k)}_{SEL})$ is computed, and $\mathcal{S}(B^{(k)}_{SEL}) \geq \mathcal{S}(A)$ is to be satisfied. It implies that classification error $\varepsilon$, should be reduced compared to an initial error with all features, $\varepsilon_A > \varepsilon_{B_{SEL}}$ and for the inclusion of each relevant feature into $B^{(k)}_{SEL}$, $\varepsilon$ is getting reduced. Next goal is to remove redundant features from $B^{(k)}_{SEL}$ and to generate an optimal subset $B^{(p)}_{OPT}$ from $B^{(k)}_{SEL}$ using backward elimination. Thus, $B^{(p)}_{OPT}$ is initialized with $B^{(k)}_{SEL}$. Based on the same order of rank, these $k$ features are used for elimination until no further feature can be removed. If removal of any feature $f_i$ retains or improves the accuracy then discard $f_i$ from $B^{(p)}_{OPT}$.

If $\mathcal{S}(B^{(p-1)}_{OPT}) \geq \mathcal{S}(B^{(p)}_{OPT})$ then, $B^{(p-1)}_{OPT} \leftarrow B_{OPT} \setminus \{\hat{f}_i\} \quad (19)$

where, $i=1,2,...,k$. After eliminating the redundant features, $\mathcal{S}(B^{(p)}_{OPT})$ is likely to be increased, $\mathcal{S}(B^{(p)}_{OPT}) \geq \mathcal{S}(B^{(k)}_{SEL})$ and $p<k$. At the end of this process, the final accuracy is $\mathcal{S}(B^{(p)}_{OPT})$ and return the optimal feature subset $B^{(p)}_{OPT}$.

A feature $f_i$ with lower discriminating ability can improve accuracy when included in $B^{(k)}_{SEL}$. On the contrary, a highly discriminative feature $f_i$ can degrade the accuracy when added in $B^{(k)}_{SEL}$. Similarly, in backward process, a feature $f_i$ with higher discriminative power can enhance the accuracy when eliminated from $B^{(p)}_{OPT}$. Nevertheless, a feature $f_i$ with lower discriminating ability can degrade the accuracy when discarded from $B^{(p)}_{OPT}$. Therefore, during backward elimination, various combinations of features can be formed and tested to reduce the redundancy which exists in $B^{(p)}_{OPT}$. Here, the same rank-based sequence of features has been maintained for deletion as followed for inclusion earlier. Furthermore, arbitrary sequences of feature combinations for elimination are also tested. In few cases, results are degraded as compared to the results reported in Table III. Again, similar accuracies with smaller deviations are obtained at the cost of redundant features. Thus, exclusion of features at any random order has not produced a further gain in accuracy. Removal of the randomly selected feature is followed mainly where feature space is very high, typically in terms of thousands or even more. Notably, feature space is limited within hundreds in this work, and sequential elimination is followed to obtain $B^{(p)}_{OPT}$.

Time complexity depends on the cardinality of input set ($A$) and an optimal subset ($B_{OPT}$), and can be defined as $O(|A|.|B_{OPT}|)$. In forward selection, one feature is included at a time in $B_{OPT}$ and tested for accuracy. Thus, the complexity of forward selection is $O(w.k)$. Backward elimination starts with $k$ features and $p<k$, thus, time complexity is $O(k^2)$, where $w$, $k$, and $p$ are defined earlier.

### C. Classifier Specification

Feature selection method and identification experiments are performed using the kNN [14] and Random Forest (RF) classifiers [30]-[31]. These two classifiers are also used for gloved hand classification from palm images in [36]. The kNN is a popular and simple supervised learning algorithm and widely used in pattern recognition. It is suitable for its lower time complexity. On the other side, the RF is a collection of decision trees with a higher accuracy of prediction for classification. It works well on a large dataset, and suitable for classification with noisy data. Every tree of the forest is independently grown to predict a particular decision on unseen test features. Tree bagging method is used for classifying a test vector based on a given training dataset. A test vector is provided for which a predicted score is calculated on the trained ensemble data. This score is the weighted average of the matching probability produced by each classification tree, and it determines the recognized class. Enhanced accuracy can be achieved with a higher number of ensemble trees. Here, the number of bagged decision trees has been varied sequentially up to 150 and beyond this range, performance degrades. Sometimes, few redundant results are produced which can be achieved with a lesser number of trees.

### D. Finger biometric optimal feature selection

To speed up the selection process, accuracies of the features are determined using the feature sets of four fingers of 100 subjects. This evaluation method can be implemented using the following approach. A feature subspace of 100 subjects is chosen randomly from the total feature set of all subjects. Then, the accuracy of each candidate attribute is determined by kNN. Here, at least five different combinations of 100 individuals are chosen and tested. Based on the average accuracy, a rank is assigned to every feature. Alternatively, a specific single finger is chosen, and accuracy of each feature is calculated. However, the main problem is redundancy in results. It implies that several features may provide the same accuracy based on this single finger-based method. Furthermore, discriminating ability of the results is low and unreliable. Hence, the first approach, based on four fingers is followed to find the $B^{(p)}_{OPT}$. The average accuracies of all the features are presented in Table II.

TABLE III
ACCURACY (%) OF FEATURE SUBSET OF RIGHT HAND (RH) AND LEFT HAND
(LH) DURING FEATURE SELECTION USING kNN BASED ON 100 SUBJECTS

| Set | Selected feature as per rank | k-NN | Forward fet. selection | k-NN | Optimal feature subset selection | k-NN |
|---|---|---|---|---|---|---|
| LH: $A_1$ | $a^{(4)}=\{a_5,a_4,a_8,a_1\}$ | 93 | $B^{(4)}_{SEL}= a^{(4)}$ | 93 | $B^{(2)}_{OPT}=\{a_5, a_4\}$ | 94 |
| LH: $A_2$ | $b^{(8)}=\{b_3,b_2,b_1,b_8,b_6,b_5,b_7,b_9\}$ | 97 | $B^{(12)}_{SEL}= B^{(4)}_{SEL}\cup b^{(8)}$ | 95 | $B^{(7)}_{OPT}= B^{(2)}_{OPT} \cup \{b_1, b_2,b_5,b_7, b_9\}$ | 97 |
| LH: $A_3$ | $c^{(9)}=\{c_5,c_7,c_8,c_9,c_4,c_3,c_1,c_2,c_{10}\}$ | 93 | $B^{(21)}_{SEL}= B^{(12)}_{SEL}\cup c^{(9)}$ | 99 | $B^{(12)}_{OPT}=B^{(7)}_{OPT} \cup \{c_1, c_2,c_4, c_8,c_{10}\}$ | 100 |
| RH: $A_1$ | $a^{(5)}=\{a_8, a_3, a_5, a_4, a_2\}$ | 92 | $B^{(5)}_{SEL}= a^{(5)}$ | 92 | $B^{(2)}_{OPT}=\{a_5,a_4\}$ | 92 |
| RH: $A_2$ | $b^{(6)}=\{b_3, b_8, b_6, b_9, b_5, b_7\}$ | 97 | $B^{(11)}_{SEL}= B^{(5)}_{SEL}\cup b^{(6)}$ | 96 | $B^{(5)}_{OPT}= B^{(2)}_{OPT} \cup \{b_3,b_9,b_7\}$ | 97 |
| RH: $A_3$ | $c^{(10)}=\{c_8,c_3,c_9,c_5,c_6,c_7,c_4,c_1,c_{10}, c_2\}$ | 95 | $B^{(21)}_{SEL}= B^{(11)}_{SEL}\cup c^{(10)}$ | 99 | $B^{(12)}_{OPT}= B^{(5)}_{OPT} \cup \{c_1,c_2,c_3,c_4,c_5,b_8,b_9\}$ | 100 |

Initially, altogether 30 features are extracted per finger, and 21 features are selected by forward selection. Afterward, eliminating some redundant features from subset $B^{(21)}_{SEL}$, only 12 features per finger are included in the $B^{(12)}_{OPT}$ and presented in Table III. Hence, at least, 60% features are neglected. Furthermore, three more features are eliminated greedily with a minimum compromise to the accuracy ($\Delta$) and it is defined as

$$\mathcal{S}(B_{OPT}\setminus\hat{f}_i) - \mathcal{S}(B_{OPT})\leq \Delta \qquad (20)$$

In the case of Eq.20, a feature $\hat{f}_i$ is removed from $B^{(12)}_{OPT}$ if its relevance in terms of accuracy gain is insignificant with respect to a permissible accuracy $\Delta$ which is determined based on the observation during backward elimination. Here, the value of $\Delta$ is chosen at most 1%. Therefore, another optimal subset $B^{(9)}_{OPT}$ is obtained based on Eq. 20. Finally, two optimal subsets are considered for experimentation. In the first and second optimal subsets, 9 ($B^{(9)}_{OPT}$) and 12 ($B^{(12)}_{OPT}$) features are selected, respectively. For the right-hand, according to initial indexing (Table II), the optimal subsets are

$B^{(9)} = \{a_4, a_5, b_1, b_5, b_9, c_1, c_4, c_8, c_{10}\}$ (21)
$B^{(12)} = \{a_4, a_5, b_1, b_2, b_5, b_7, b_9, c_1, c_2, c_4, c_8, c_{10}\}$ (22)

For the left-hand, most of the features are same as right-hand.
$B^{(9)} = \{a_4, a_5, b_9, c_1, c_2, c_3, c_4, c_5, c_9\}$ (23)
$B^{(12)} = \{a_4, a_5, b_3, b_7, b_9, c_1, c_2, c_3, c_4, c_5, c_8, c_9\}$ (24)

The optimal subsets of selected features are given in Table III. It is evident that an optimal feature subspace should not necessarily to be unique. Optimality in feature selection may change according to dependency on the target class, which is determined by finding the correlation among the attributes during the training phase.

### E. Feature overfitting issues and validation

Feature overfitting is one major issue in supervised classification that performs well on training dataset; however, performance degrades during testing on unknown data. To avoid overfitting problem and to validate attribute selection method, 10-fold cross validation (CV) [23] scheme using kNN ($k=$ 1 to 5) has been performed. It is a popular method for selecting good attributes at the highest relevance with the target class. Here, feature relevance is characterized by the correlation [23]. The linear Pearson correlation coefficient [28], and the Euclidean distance (ED) are used for 10-fold CV purpose. The linear Pearson correlation coefficient (PC) is invariant to translation and rotation and defined as

$$PC(U_j,V)= \frac{(U_j-\overline{U}_j)(V-\overline{V})}{\sqrt{(U_j-\overline{U}_j)^2(V-\overline{V})^2}} \qquad (25)$$

where, $U_j$ is the $j^{th}$ feature of vector $U$, and $V$ is the class label and, $\overline{U}$ and $\overline{V}$ are the respective mean; $U$ and $V$ are chosen to be random variables. The bagging method using randomly grown ensemble of decision trees of the RF classifier has also been used for this validation purpose. It computes the unbiased average out-of-bag (OOB) error on the true estimated prediction used for training. Generally, 90% samples are randomly selected and trained, while remaining 10% feature vectors are validated. Initially, altogether 300 features from 100 subjects are chosen, out of which 270 random feature vectors are trained, and validated by remaining 30 feature vectors at each fold. Similarly, feature vectors from 200 subjects are considered randomly. In this case, 540 samples are trained and validated by the rest 60 samples using both of the classifiers. Average 10-fold CV errors using kNN are given in Table IV. As the number of nearest neighbors increases from $k$=1 to $k$=5, the CV error rates also increase. Interestingly, at $k$=1, the CV error rates are zeroes for all the subject spaces of any hand using either of the distance metrics. However, for other values of $k$, the errors calculated by the Pearson correlation are higher than the Euclidean distance. It is obvious that $B^{(12)}_{OPT}$ provides better accuracies than $B^{(9)}_{OPT}$. The OOB errors are given in Table V. The classification trees are varied from RF=50 to RF=150. The OOB errors decrease with higher values of RF. Particularly, from RF=130 to RF=150, average OOB error is 0.0346 for both of the hands of 100 subjects using $B^{(12)}_{OPT}$. Average CV errors and OOB errors using RF are reasonable.

### IV. EXPERIMENTAL DESCRIPTIONS AND RESULTS

A well-known hand dataset containing more that 600 subjects with sufficient posture variations is preferred here to determine inter-class variability. However, other datasets (e.g., IITD, CASIA, etc.) contain a lesser number of subjects.

### A. Database Specification

The Bosphorus hand database was created at the BOGAZICI University [6]. The *HandGeometryDBPart1* is its larger sub-dataset that contains the left- and right-hand images of 642 individuals, who are the French and Turkish students and staff members from different universities, and age limit is between 20 and 50. Three images per hand have been collected at by a HP Scanjet scanner with 383×526 pixels at 45 dpi, at three different sessions. Time lapse is varied from 2 weeks to 3 years, and average time lapse is one year. Experiments have been conducted on both types of hand images of 638 subjects. Four subjects are excluded (*see* Fig.11), at least one per subject are not suitable for preprocessing.

TABLE IV
10-FOLD CROSS VALIDATION ERROR USING kNN CLASSIFIER DURING FEATURE SELECTION BASED ON THE 100 AND 200 SUBJECTS. FOR 100 SUBJECTS: TRAINING SET SIZE: 270 AND TEST SET SIZE: 30. FOR 200 SUBJECTS: TRAINING SET SIZE: 540 AND TEST SET SIZE: 60

| Subject | Metric | 9 features per finger | | | | | 12 features per finger | | | | |
|---|---|---|---|---|---|---|---|---|---|---|---|
| | | k=1 | k=2 | k=3 | k=4 | k=5 | k=1 | k=2 | k=3 | k=4 | k=5 |
| RH: 100 | PC | 0 | 0.07 | 0.083 | 0.113 | 0.153 | 0 | 0.056 | 0.063 | 0.10 | 0.11 |
| RH: 100 | ED | 0 | 0.03 | 0.04 | 0.08 | 0.116 | 0 | 0.023 | 0.033 | 0.07 | 0.09 |
| RH: 200 | PC | 0 | 0.088 | 0.116 | 0.148 | 0.183 | 0 | 0.086 | 0.10 | 0.135 | 0.166 |
| RH: 200 | ED | 0 | 0.045 | 0.058 | 0.116 | 0.138 | 0 | 0.043 | 0.05 | 0.093 | 0.11 |
| LH: 100 | PC | 0 | 0.093 | 0.106 | 0.17 | 0.22 | 0 | 0.043 | 0.06 | 0.113 | 0.143 |
| LH: 100 | ED | 0 | 0.06 | 0.076 | 0.116 | 0.153 | 0 | 0.033 | 0.037 | 0.073 | 0.087 |
| LH: 200 | PC | 0 | 0.123 | 0.15 | 0.198 | 0.231 | 0 | 0.063 | 0.072 | 0.115 | 0.157 |
| LH: 200 | ED | 0 | 0.085 | 0.108 | 0.171 | 0.208 | 0 | 0.053 | 0.036 | 0.087 | 0.127 |

TABLE V
OUT-OF-BAG CLASSIFICATION ERROR USING RF CLASSIFIER DURING FEATURE SELECTION BASED ON THE 100 AND 200 SUBJECTS

| RF | RH: 100 Subj. | | RH: 200 Subj. | | LH: 100 Subj. | | LH:200 Subj. | |
|---|---|---|---|---|---|---|---|---|
| | $B^{(9)}$ | $B^{(12)}$ | $B^{(9)}$ | $B^{(12)}$ | $B^{(9)}$ | $B^{(12)}$ | $B^{(9)}$ | $B^{(12)}$ |
| 50 | 0.12 | 0.10 | 0.17 | 0.14 | 0.12 | 0.09 | 0.17 | 0.145 |
| 60 | 0.10 | 0.09 | 0.15 | 0.13 | 0.11 | 0.076 | 0.16 | 0.133 |
| 70 | 0.09 | 0.086 | 0.13 | 0.125 | 0.10 | 0.063 | 0.151 | 0.131 |
| 80 | 0.083 | 0.076 | 0.128 | 0.111 | 0.096 | 0.05 | 0.143 | 0.111 |
| 90 | 0.08 | 0.063 | 0.111 | 0.105 | 0.09 | 0.046 | 0.12 | 0.11 |
| 100 | 0.07 | 0.056 | 0.10 | 0.088 | 0.083 | 0.04 | 0.116 | 0.10 |
| 110 | 0.067 | 0.056 | 0.091 | 0.088 | 0.08 | 0.036 | 0.111 | 0.096 |
| 120 | 0.06 | 0.05 | 0.081 | 0.082 | 0.073 | 0.033 | 0.108 | 0.095 |
| 130 | 0.05 | 0.046 | 0.078 | 0.08 | 0.07 | 0.026 | 0.101 | 0.093 |
| 140 | 0.046 | 0.043 | 0.076 | 0.076 | 0.06 | 0.03 | 0.095 | 0.086 |
| 150 | 0.043 | 0.033 | 0.07 | 0.07 | 0.063 | 0.03 | 0.098 | 0.073 |

TABLE VI
ACCURACY (%) OF EVERY SUBJECT PARTITION USING kNN CLASSIFIER, CONSIDERING ALL THE FINGERS WITH $B^{(12)}$

| $T_{CASE}$ | $t_1$ | $t_2$ | $t_3$ | $t_4$ | $t_5$ | $t_6$ | $t_7$ | $t_8$ | $t_9$ | $t_{10}$ | $t_{11}$ | $t_{12}$ | $t_{13}$ |
|---|---|---|---|---|---|---|---|---|---|---|---|---|---|
| RH:$T_{C-1}$ | 96 | 96 | 92 | 90 | 98 | 96 | 96 | 96 | 100 | 92 | 98 | 96 | 73.7 |
| RH:$T_{C-2}$ | 97.4 | 96 | 90 | 90 | 98 | 96 | 96 | 96 | 100 | 92 | 98 | 96 | 76 |
| LH:$T_{C-1}$ | 98 | 100 | 96 | 94 | 88 | 92 | 98 | 98 | 98 | 94 | 100 | 96 | 97.4 |
| LH:$T_{C-2}$ | 100 | 98 | 100 | 96 | 92 | 86 | 98 | 98 | 100 | 90 | 100 | 94 | 98 |

### B. Identification

In identification, matching is performed between the test feature set and all trained feature templates. As only three images per hand had been acquired, two images are chosen for training, and one is for testing. Altering the images for training and testing, three different combinations are formulated, by considering every unique image as a test sample one at a time and remaining two samples for training. Every combination is experimented independently and average accuracy of those executions is calculated for different populations. Total 638 individuals are divided into seven different subsets of 100, 200, 300, 400, 500, 600, and 638 subjects based on random combinations. Though, this division is trivial for calculating the accuracy when all the subjects are considered. However, when the performances from 100 to 600 individuals are evaluated, it plays a significant role in such assessment. Results may vary for a different combination of randomly chosen subjects for a given sample size. The selected feature matrix (T) of 600 subjects is divided into 12 segments from $t_1$ to $t_{12}$, and 50 persons are considered in each segment equally. The last segment $t^*_{13}$ contains the features of remaining 38 subjects. All the segments are labeled from $t_1$ to $t_{13}$. This particular partitioning of subjects is denoted by $T_{C1}$. Alternatively, another partitioning has been denoted as $T_{C2}$; where $t^*_1$ contains first 38 subjects and remaining each of $t_2$ to $t_{13}$ segment contains 50 subjects. The advantage of partitioning the subjects is that different combinations can be formed with various segments from $t_1$ to $t_{13}$. As an example, for 100 subjects, considering all equal-size partitions, $[t\times(t-1)/2]$ =66 different combinations are formed and for each case, three unique test sets are experimented. Thus, total 66×3=198 tests are conducted and the average results are reported. This method is continued for all the population stages from 200 to 600. The accuracy of each partition using kNN is given in Table VI.

$$T_{C1}= \{t_1, t_2, t_3, t_4, t_5, t_6, t_7, t_8, t_9, t_{10}, t_{11}, t_{12}, t^*_{13}\} \quad (26)$$
$$T_{C2}= \{t^*_1, t_2, t_3, t_4, t_5, t_6, t_7, t_8, t_9, t_{10}, t_{11}, t_{12}, t_{13}\} \quad (27)$$

It is noted that $T_{C1}$ has produced better results than $T_{C2}$, and the difference is ±2% approximately. Average accuracies of every partition except the smallest one of $T_{C1}$ and $T_{C2}$ for any hand are more than 95.1%.

In various experiments, the subsets $B^{(9)}$ and $B^{(12)}$ of every finger are tested by kNN and RF. This finger-level testing has been performed to calculate the uniqueness of each finger. Every finger is arranged according to its discriminative correctness using the majority of voting, and a rank is assigned correspondingly. For this purpose, accuracies of each finger for 600 and 638 subjects are used, reported in Table VII. In this experiment, the fingers of right-hand provide better accuracies than left-hand. Rank assignment to every finger is crucial to conduct other experiments. For example, the middle finger of the right-hand with $B^{(9)}$ has produced the highest accuracy. Thus, features of this finger are considered initially, and its accuracy is evaluated for all population subdivisions. After that, features of the ring finger are included with the middle finger and tested the accuracy. This process is followed by other fingers. However, in other cases, the ring finger is significantly more relevant than the middle finger. In the case of $B^{(12)}$, fingers are arranged according to their order of performance. Ring finger is tested first, and then the middle finger is combined with it. Other fingers are included in the same fashion and each subpopulation is tested. Though, the first choice of selecting a finger i.e., either the middle finger in $B^{(9)}$ or the ring finger in $B^{(12)}$ is not important because these two fingers are combined at the next step. For any population, several tests are executed for every combination of subjects which are obtained based on partitioning as mentioned earlier. Multiple executions of different sets of subjects are conducted using RF to verify the reliability of accuracies. The minimum numbers of classification trees to achieve the results in these present experimental scenarios are also reported within parenthesis in Table VIII - IX. Finally, four fingers and five fingers are combined for producing a better identity of an individual. Results imply that while the thumb is included with other four fingers, the overall performance degrades. As thumb is the most flexible finger, its features calculation suffers mostly due to inter-finger spacing status. Experiments including the thumb with other fingers cannot provide good results in all situations. Thus, the four fingers are preferred in some existing works. This work investigates the accuracies of both four and five fingers. This system produces the best results using four fingers. The difference between the accuracies of four and five fingers is approximately ±3%. The results also indicate how a specific finger is significant to determine the identity of a per-son. The differences of accuracies between four fingers of the left- and right-hand subjects are nearly ±2% using RF.

TABLE VII
IDENTIFICATION ACCURACY (%) OF EVERY FINGER

| Hand | Subjects | Classifier | B⁽⁹⁾ : 9 features per finger | | | | | B⁽¹²⁾ : 12 features per finger | | | | |
|---|---|---|---|---|---|---|---|---|---|---|---|---|
| | | | Little | Ring | Middle | Index | Thumb | Little | Ring | Middle | Index | Thumb |
| Right | 600 | kNN | 36.2 | 67.5 | 68.5 | 53.2 | 28.9 | 45.5 | 72 | 68.2 | 61 | 32.4 |
| | | RF | 42.84 | 69.9 | 70.7 | 62.5 | 34.84 | 48.4 | 74.5 | 72.5 | 63.7 | 38.2 |
| | 638 | kNN | 34.2 | 64.3 | 65.9 | 50.8 | 27.5 | 43.9 | 69.9 | 66.4 | 59.3 | 31.1 |
| | | RF | 40.8 | 67.1 | 68.4 | 60.5 | 32.92 | 47.4 | 73.5 | 71.1 | 62.4 | 36.7 |
| Left | 600 | kNN | 32.17 | 59.0 | 55.67 | 49.17 | 26.84 | 40.67 | 68.5 | 64.5 | 59.34 | 28.67 |
| | | RF | 39.34 | 61.17 | 56.83 | 53.84 | 26 | 49.34 | 71.84 | 67.67 | 63.84 | 33 |
| | 638 | kNN | 30.73 | 57.84 | 54.1 | 47.5 | 23.84 | 39.7 | 67.25 | 62.7 | 57.4 | 27.28 |
| | | RF | 38.01 | 58.78 | 56.43 | 52.1 | 24.77 | 46.4 | 70.22 | 64.58 | 64.42 | 31.2 |

TABLE VIII
IDENTIFICATION ACCURACY (%) USING COMBINATION OF THE RIGHT-HAND FINGERS

| Subjects | Classfr. | B⁽⁹⁾ : 9 features per finger | | | | | B⁽¹²⁾ : 12 features per finger | | | | |
|---|---|---|---|---|---|---|---|---|---|---|---|
| | | Little | Ring | Middle | Index | Thumb | Little | Ring | Middle | Index | Thumb |
| 100 | kNN | 88 | 99 | 99 | 100 | 98 | 90 | 98 | 100 | 100 | 98.5 |
| | RF | 97(105) | 99(59) | 100(49) | 100(42) | 100(50) | 92(99) | 99(86) | 100(53) | 100(60) | 100(64) |
| 200 | kNN | 78.5 | 94.5 | 98.5 | 99 | 96.5 | 84 | 96 | 100 | 100 | 97.5 |
| | RF | 84.5(115) | 95.5(110) | 98(104) | 99.5(101) | 99(123) | 84(79) | 98(105) | 100(139) | 100(81) | 99.5(90) |
| 300 | kNN | 78.67 | 93.34 | 97 | 97.67 | 95 | 80.34 | 94 | 98.34 | 98 | 95.67 |
| | RF | 83.34(103) | 93.34(145) | 97.67(132) | 98.34(108) | 97.67(128) | 80.67(99) | 95(123) | 97.67(144) | 99(131) | 98.34(126) |
| 400 | kNN | 77.25 | 89.5 | 95.5 | 96.25 | 93.75 | 77.25 | 91.75 | 96.75 | 96.5 | 94.25 |
| | RF | 80(130) | 91.5(130) | 96.5(127) | 97.5(148) | 96(121) | 77.5(126) | 92.75(114) | 96(112) | 98.5(111) | 97(127) |
| 500 | kNN | 71.6 | 87.8 | 93.2 | 94.8 | 91.6 | 74.6 | 89.8 | 96 | 95.6 | 92.6 |
| | RF | 75.4(128) | 89.8(135) | 94.8(129) | 97(147) | 95.8(149) | 75.8(144) | 91.2(111) | 95.2(143) | 98.2(141) | 96.4(99) |
| 600 | kNN | 68.5 | 85.84 | 92 | 92.5 | 89.84 | 72 | 88 | 94.17 | 94.5 | 91.5 |
| | RF | 69.84(124) | 86.92(128) | 93.87(144) | 94.34(129) | 93.2(137) | 74.17(150) | 89.67(147) | 94.17(130) | 97.5(135) | 96.17(133) |
| 638 | kNN | 65.83 | 84.39 | 90.82 | 91.29 | 88.78 | 69.91 | 85.27 | 91.54 | 92.64 | 89.35 |
| | RF | 68.81(120) | 85.87(133) | 91.93(141) | 92.95(136) | 91.7(138) | 70.9(125) | 87.94(128) | 92.1(150) | 96.56(145) | 94.68(142) |

TABLE IX
IDENTIFICATION ACCURACY (%) USING COMBINATION OF THE LEFT-HAND FINGERS

| Subjects | Classfr. | B⁽⁹⁾ : 9 features per finger | | | | | B⁽¹²⁾ : 12 features per finger | | | | |
|---|---|---|---|---|---|---|---|---|---|---|---|
| | | Little | Ring | Middle | Index | Thumb | Little | Ring | Middle | Index | Thumb |
| 100 | kNN | 79 | 95 | 99 | 100 | 100 | 90 | 98 | 100 | 100 | 99 |
| | RF | 82(124) | 97(49) | 99(66) | 100(73) | 100(98) | 94(98) | 99(79) | 100(41) | 100(36) | 100(30) |
| 200 | kNN | 75.5 | 86.5 | 96.5 | 99 | 99 | 83 | 93 | 98.5 | 100 | 100 |
| | RF | 75(125) | 91.5(148) | 97(132) | 98.5(147) | 96.5(128) | 88.5(116) | 97(100) | 99(115) | 99.5(74) | 98.5(103) |
| 300 | kNN | 69.34 | 84 | 93.34 | 98 | 94.67 | 75.34 | 89.67 | 95.67 | 99 | 97.34 |
| | RF | 72(142) | 90(141) | 94.67(116) | 97(123) | 96.34(125) | 78(101) | 91(111) | 96.67(139) | 98.34(142) | 97.67(149) |
| 400 | kNN | 65 | 83.25 | 93.5 | 96.5 | 94 | 73.25 | 88.25 | 95.5 | 98.25 | 95.25 |
| | RF | 65.75(124) | 87.25(126) | 94.25(145) | 96.25(138) | 96(135) | 77.25(145) | 91(139) | 95.25(140) | 97.75(135) | 96.75(148) |
| 500 | kNN | 61.8 | 80.6 | 92.4 | 95 | 93.4 | 70.8 | 87.4 | 95.2 | 97 | 93.84 |
| | RF | 63(134) | 84.2(140) | 92.6(135) | 94.8(121) | 95(126) | 75(127) | 90.2(143) | 95.2(120) | 97.4(136) | 96.6(148) |
| 600 | kNN | 59 | 76.84 | 89.5 | 92.17 | 91.34 | 68.5 | 84.17 | 92.84 | 94.84 | 92.84 |
| | RF | 59.34(132) | 81.5(126) | 90.84(147) | 93(142) | 93.34(149) | 71(132) | 87.17(148) | 93.34(127) | 96.67(143) | 93.5(139) |
| 638 | kNN | 57.84 | 75.24 | 88.41 | 91.23 | 90.29 | 67.25 | 82.45 | 91.54 | 93.26 | 90.76 |
| | RF | 58.16(148) | 81.04(146) | 90.29(143) | 92.32(147) | 91.7(145) | 69.75(145) | 85.27(122) | 92.8(133) | 95.92(145) | 92.79(148) |

Identification accuracies of a subject and that is chosen randomly and labeled as the 100th subject, are presented in Fig.4 and Fig.5. The OOB classification error estimation by RF is also important. It computes the average of cumulative misclassification probability for OOB observations in the bootstrap dataset, shown in Fig.4. About one-third of data is left out from tree construction at any stage, and considered as OOB data which is used for testing. The OOB error increases with a higher population using a fixed number of classification trees and decreases with a higher number of classification trees for a specific subject space. The predicted matching score and its standard deviation of a legitimate user should be higher compared to other imposter classes for identification. In this regard, an example is illustrated in Fig.5(a-b), to differentiate between the predicted matching scores of the 100th legitimate subject and the set of imposter subjects for correct identification. Fig.6 shows how the accuracies have been measured based on the Euclidean distances between the feature vectors using kNN. For this purpose, the set of the right-hand subjects is experimented using $B^{(12)}$. Lesser Euclidean distance indicates more accurate intra-class similarities.

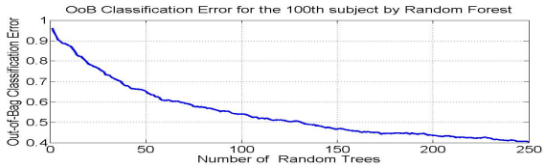

Fig.4. OOB classification error estimation of the 100th subject using $B^{(12)}$ of four fingers. It is the generalization error estimated on the training set by RF.

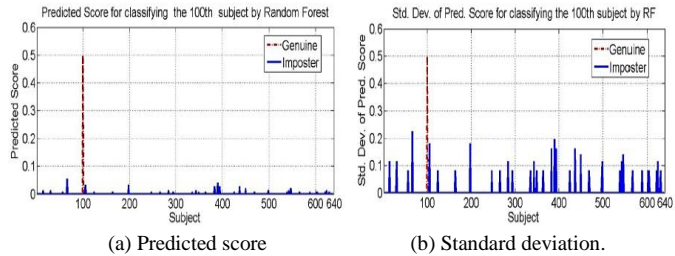

(a) Predicted score    (b) Standard deviation.

Fig.5. Identification of the 100th right-hand subject using $B^{(12)}$ of four fingers by RF. The predicted score is the average of underlying class probabilities of the OOB observations.

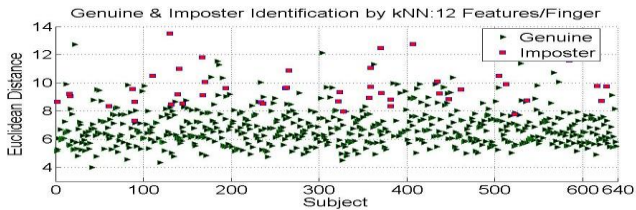

Fig.6. Identification of right-hand subjects, shown using scatter plot.

The genuine and imposter represent the correct and wrong identification of subjects, respectively. The mean Euclidean distance is 6.81 with the standard deviation (*std*) of 1.29 for correct identification of right-hand subjects using $B^{(12)}$. Fig.7 represents the accuracies measured by varying classification trees from 100 to 150 for different subsets of individuals using $B^{(12)}$. The best results achieved for all population subsets are marked and connected by a dotted line for better clarity. For this purpose, different combinations are tested for each population division with 1-10 weeks of the time interval for observing the randomness of classification trees. A different bootstrap dataset is envisaged for every execution. In some cases, it is observed that the results are very similar. Overlapping of data points indicates the extent of accuracy measured for various subpopulations. Every combination of subjects is executed at least ten times, and RF provides excellent results for correct identification with about ±3% to ±4% randomness in the predicted scores.

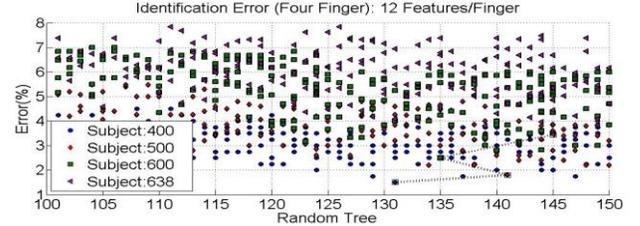

Fig.7. Identification error (%) of various subpopulation using the right-hand $B^{(12)}$ using RF is shown using scatter plot. Overlapped data markers indicate same results of different executions.

### C. Verification

Verification is experimented to differentiate a legitimate subject from an imposter subject. The performance is computed regarding the EER. The EER is defined as the point at which values of the FAR and FRR are same. In verification, a claimed feature vector is compared to his/her stored templates based on a given distance threshold. Distances between a claimer and all the enrolled feature set are calculated. If the distances are within the threshold, then the person is accepted as genuine; otherwise rejected as an imposter. Here, the standardized Euclidean distance is applied and defined as

$$\Re_{i,g} = \left[ \sum_{j=1}^{\omega} \frac{(\alpha_{i,j} - \beta_{g,j})^2}{\sigma_j^2} \right]^{1/2} \quad (28)$$

where, $\alpha_{i,j}$ represents the $j^{th}$ feature of the $\alpha_i$ enrolled user; $\beta_{g,j}$ denotes the $j^{th}$ test feature of the claimer $\beta_g$; $\sigma_j$ is the standard deviation of the $j^{th}$ feature, and $\omega$ is the number of selected features. The initial distance threshold is determined by the mean of minimum distances of the enrolled subjects, and it is varied accordingly to calculate the EER. The population is considered as a blend up of genuine and imposter feature vectors. For every combination of genuine users ($N_G$), two samples ($N_{tr}$=2) are used for training, and the remaining one sample ($N_{ts}$=1) is tested for verification. The required parameters are defined in the context of $N_G$=638 subjects. Training set=$N_{tr} \times N_G$=1276; Test set=$N_{ts} \times N_G$=638; Genuine comparison=$N_{tr} \times N_G \times N_{ts}$=1276; Imposter comparison =$N_{tr} \times (N_G-1) \times N_G$=812812; and Total comparison=Genuine comparison + Imposter comparison= 814088. Various populations are considered for conducting experiments using both the sets of selected features of four fingers. Three combinations are considered for each population and experimented. The average EER is reported in Table X. The EERs of left- and right-hands for a higher population with 600 and 638 subjects are very similar. The Receiver Operating Characteristic (ROC)[4] curves of every subpopulation of the both hands using $B^{(9)}$ and $B^{(12)}$ are represented in Fig.8-9.

TABLE X
VERIFICATION PERFORMANCE USING FOUR FINGERS

| Hand | Selected Features | EER of various population | | | | |
|---|---|---|---|---|---|---|
| | | 400 | 458 | 500 | 600 | 638 |
| Right | $B^{(9)}$ | 0.0660 | 0.070 | 0.069 | 0.072 | 0.080 |
| | $B^{(12)}$ | 0.0670 | 0.070 | 0.072 | 0.074 | 0.078 |
| Left | $B^{(9)}$ | 0.0692 | 0.0685 | 0.0660 | 0.0778 | 0.0823 |
| | $B^{(12)}$ | 0.0623 | 0.0614 | 0.0620 | 0.0758 | 0.0783 |

TABLE XI
VERIFICATION PERFORMANCE OF DISJOINT SUBJECTS USING $B^{(12)}$ OF FOUR FINGERS

| Genuine subjects | 400 | 500 | 600 |
|---|---|---|---|
| Imposter subjects | 100 | 100 | 38 |
| Genuine comparison | 400×2 | 500×2 | 600×2 |
| Imposter test set | 399+300 | 499+300 | 599+114 |
| Imposter comparison | 800×699 | 1000×799 | 1200×713 |
| Total comparison | 800×700 | 1000×800 | 1200×714 |
| Right hand EER | 0.0700 | 0.0750 | 0.0755 |
| Left hand EER | 0.0625 | 0.0635 | 0.0741 |

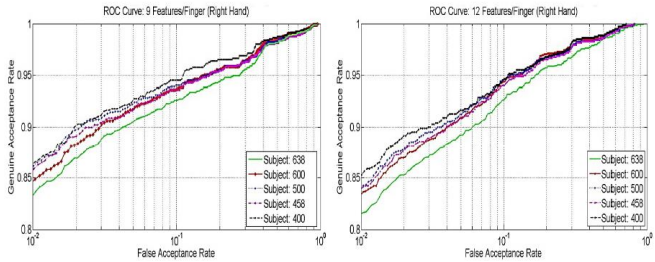

Fig.8. ROC curves of right-hands using subsets $B^{(9)}$ and $B^{(12)}$ of four fingers.

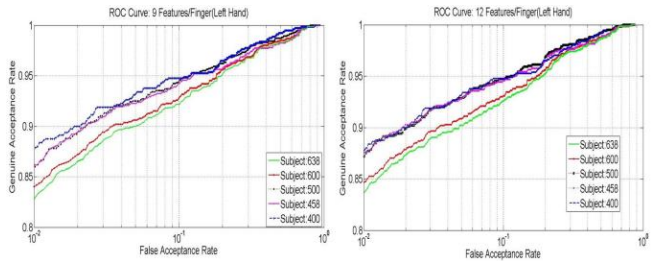

Fig.9. ROC curves of the left-hands using subsets $B^{(9)}$ and $B^{(12)}$ of four fingers.

In another verification scheme, firstly, 400 legitimate subjects are trained. Also, another unknown 100 subjects are considered as imposter set. The genuine and imposter feature sets are chosen to be from completely disjoint subject spaces. The imposter set contains 300 feature vectors which are included with 400 genuine test set. Finally, it creates a mixed feature space of 700 templates, out of which only one vector is genuine for every trained subject. After that, verification is performed using $B^{(12)}$ only. Similar experiments have been carried out for other two subpopulations with 500 and 600 subjects. The defined parameters and EERs are reported in Table XI; and the ROC curves are presented in Fig.10.

### D. Performance Comparison

The performances of the proposed work are compared with other existing works on the Bosphorus hand database. Mainly, those existing works are compared which are experimented with 400 to 638 subjects on this database and provided in Table XII. Although the listed works have been experimented with the same hand images per subject, the methods of the preprocessing, feature extraction, classification algorithm, and other important parameters are not identical. The comparison is provided for completeness of the study made in this paper. In this work, both hands are used for experimental comparison with other existing works. Results of the proposed work using four fingers with $B^{(12)}$ are mentioned. To compare with [6], experiments have been conducted with the right-hand images of 458 subjects. Identification accuracies achieved using kNN and RF classifiers are 96.3% and 98.04% (min. RF: 143), respectively. Similarly, for the left-hand images of 458 subjects, accuracies are 97.6% and 96.95% (min. RF: 137), using kNN and RF classifiers, respectively. However, other experiments for 458 subjects at specific finger-level based on $B^{(9)}$ or $B^{(12)}$ have not been evaluated. In [2], the results are obtained using shape-based features of two different population sets of 400 and 600 left-hand subjects. In [7], various experiments for verification, at score-level fusion are performed on the left-hand images of 638 subjects, and the minimum EER is 0.0369. It is noted that no identification result is reported in [7]. However, the results of other experimental scenarios without fusion are comparable to this proposed work. The size of shape-based feature set like ICA [2] or ART_shape [3] is typically 200, which is comparatively higher than the set of geometric features. Thus, it can be stated that for a higher magnitude of subjects, without any fusion strategy the present method based on the four fingers is competitive over other related works.

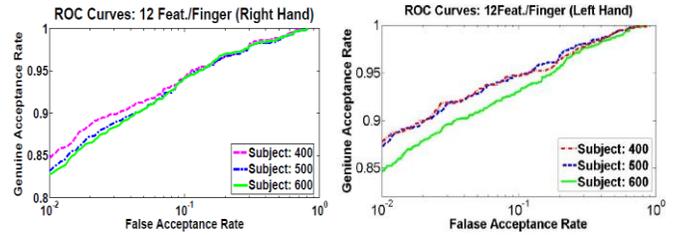

Fig.10. ROC curves of the disjoint population using $B^{(12)}$ of four fingers of right and left hands.

### E. Other Observations and Limitations

Sufficient posture variations create major challenges during normalization in the peg-free system. Here, four subjects are found unacceptable for enrolment due to flawed hand segmentation at preprocessing. However, it carries a trivial influence on the experimental results. During image acquisition, it is necessary to place fingers apart without any restrictions. However, one raw image could not be captured with complete hand which produces partial contour of the little finger, shown in Fig.11(a). The normalization method cannot integrate a larger missing shape profile of an incomplete contour. Hence, the subject is discarded to avoid additional preprocessing step for profile retrieval from an incomplete hand shape. The image, shown in Fig.11(b) cannot be preprocessed properly as the lower portion of the palm and wrist cannot be segmented in the binary image, resulting in an incomplete hand contour. Thus, it may cause incomplete profile extraction and hence, the subject is excluded. Sometimes, the nail becomes a critical factor to segment a good profile when it is associated with actual finger shape. It may cause incorrect recognition because of wrong intra-class comparison. Thus, the challenging situation for nail is neglected by excluding the subject after realizing the segmentation difficulty, shown in Fig.11(c). Thus, additional carefulness is required at image acquisition.

TABLE XII
IDENTIFICATION AND VERIFICATION PERFORMANCE COMPARISON WITH OTHER STATE-OF-THE-ART METHODS

| Authors, Year | Feature set | Subjects; Hand | Identification (%) [Verification] | Proposed, four finger geometry using $B^{(12)}$ |
|---|---|---|---|---|
| Yörük et al., 2006; [6] | Hand shape (ICA2) | 458; Right | 97.31; [GAR: 98.21] | 98.04; [EER: 0.0700] |
| A.Ei-Sallam et al., 2011; [12] | MHD and ICA | 500; Right | 98.2; [GAR: 98.5] | 98.2; [EER: 0.0720] |
| Yörük et al., 2006; [3] | ART_shape | 458; Left | 95.78; [Not Reported] | 97.6; [EER: 0.0614] |
| Dutağaci et al., 2008; [2] | Geometric with LDA | 400; Left | 97.79; [Not Reported] | 98.25; [EER: 0.0623] |
|  | PCA on binary hand | 600; Left | 95.61; [Not Reported] | 96.67; [EER: 0.0758] |
| Kang and Wu, 2014; [7] | CAF-FDs and finger area of four fingers | 638; Left | [Not Reported]; [hand shape EER: 0.1049] | 95.92; [EER: 0.0783] |

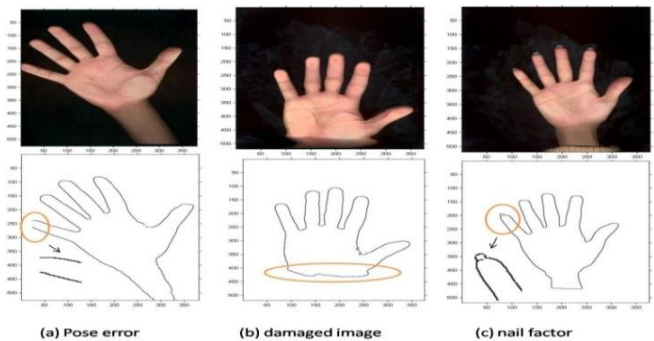

Fig.11. Three different flawed scenarios are encountered during hand segmentation. (a) whole hand area is not considered during image acquisition, (b) incomplete hand contour due to damaged raw image, (c) hand segmentation error due to the nail factor.

Rather than considering all the computed features, selection of good features improves the performance, i.e., feature selection provides better results compared to without selection. For example, identification accuracies of the right- and left-hands of 638 subjects have been improved to 10% and 11.76% by kNN for four fingers using $B^{(12)}$, respectively. Moreover, feature relevance of each subset for the right-hand images of 638 subjects i.e., $\{a_5, a_4\}$, $\{b_1, b_2, b_5, b_7, b_9\}$, $\{c_1, c_2, c_4, c_8, c_{10}\}$ have been measured by kNN as 69.34% (*std* = ±0.59%); $B^{(2)}_{OPT}$ =69.6%, and $B^{(5)}_{OPT}$ =85.6%.

Mainly, the limitations of the present method are:
 a) Unable to figure out an incomplete contour automatically.
 b) Nail-effect cannot be eradicated completely. No specific operation has been followed for eliminating the consequences of especially the nail.
 c) An isthmus on the finger caused by the ring is an important challenge during hand segmentation. Though, smaller cavities are resolved by morphological bridging and filling the holes thereof. Due to ring artifact, a finger can be completely disconnected from the main hand component and appeared as other distorted smaller component(s). If this gap is larger, i.e., the distance between the main component and disconnected finger component is little large, no additional operation has been applied to reconnect the isolated finger.
 d) As the extreme open-ended points, i.e., valleys of every finger are sensitive to noise and pose variations. Therefore, features computed bases on these points may not be completely accurate. However, those noisy features should be rejected by the feature selection algorithm.
Thus, these issues can be solved as the future enhancement. of this work.

## V. CONCLUSION

A new finger biometric system based on four fingers has been presented. The essential steps required for preprocessing are discussed. The proposed normalization algorithm follows simple steps, mainly arithmetic and logical operations are required and its computational difficulty is lesser. The complexities for both of the normalization and feature selection are in the quadratic order. This hand normalization method also supports to define shape-based features, rather than conventional geometric features of fingers. However, the normalization method can be improved further by solving the mentioned limitations. An adaptive greedy algorithm has been applied to select two subsets of highly discriminative characteristics from every normalized finger. Selection of relevant features decreases 60% of the initial feature space using $B^{(12)}$ which effectively reduces the computational time for performance evaluation which is an essential factor for online based authentication. Removal of redundant features improves the accuracies for accurate individualization. The global feature subset selection algorithm aims to reduce the classification error.

Our objective is to consider a database with larger population rather than a smaller database with more sample images per subject. Even though the present system can also be experimented with other freely available or nonproprietary databases such as the IITD, CASIA and with more challenging imaging conditions, and the performances can be compared with the related state-of-the-art methods. Results of the left- and right-hand have been compared for better clarity. In the case of four fingers with $B^{(12)}$, the differences of identification accuracies between the left- and right-hands are about 2%. For a higher population, mainly score-level fusion has been adopted to enhance the accuracy in literature. Therefore, fusion strategy at various levels such as the feature-level, score-level, and/or decision-level of both hands can be explored for enhancing accuracy. Here, the performances have been evaluated by kNN and RF classifiers. However, any other classifier like, SVM can also be used. Moreover, shape-based features using invariant shape descriptors can be extracted and tested. The magnitude of the population can be increased to raise real-world applicability of this flexible posed four-finger biometric system. Experimental results indicate that the proposed system can provide improved performance for applications with 600 or even more subjects.

## REFERENCES

[1] A.K. Jain, A. Ross, and S. Prabhakar, "An Introduction to Biometric Recognition," *IEEE Trans. on Circuits and System for Video Technology*, vol. 14, no. 1, pp. 4-20, Jan. 2004.

[2] H. Dutağaci, B. Sankur, and E.Yörük, "A comparative analysis of global hand appearance-based person recognition," *Jrnl. of Electronic Imaging*. vol. 17, no.1, pp. 011018/1–19, Jan-Mar. 2008.

[3] E.Yörük, H. Dutağaci, and B. Sankur, "Hand biometrics," *Image and Vission Computing*, vol.24, pp. 483-497, 2006.

[4] N. Duta, "A survey of biometric technology based on hand shape," *Pattern Recognition*, vol. 42, pp. 2797–2806, 2009.

[5] V. Kanhangad, A. Kumar, and D. Zhang, "A Unified Framework for Contactless Hand Verification," *IEEE Trans. on Information Forensics and Security*, vol.6, no.3, pp.1014-27, Sept. 2011.

[6] E.Yörük, E. Konukoğlu, B. Sankur, and J. Darbon, "Shape-based hand recognition," *IEEE Trans. on Image Processing*, vol.15, no.7, pp. 1803–15, 2006.

[7] W. Kang and Q. Wu, "Pose-Invariant Hand Shape Recognition Based on Finger Geometry," *IEEE Trans. on Systems, Man, and Cybernetics: Systems*, vol. 44, no. 11, pp.1510-1521, Nov. 2014.

[8] V. Kanhangad, A. Kumar, and D. Zhang, "Contactless and Pose Invariant Biometric Identification using Hand Surface," *IEEE Trans. on Image Processing*, vol. 20, no. 5, pp. 1415-1424, 2011.

[9] R.S.Reillo, C.S.Avila, and A.G.Macros, "Biometric identification through hand geometry measurements," *IEEE Trans. on Pattern Anal. Mach. Intel.*, vol. 22, no.10, pp. 1168–71, 2000.

[10] N.Charfi, H.Trichili, A.M. Alimi, and B. Solaiman, "Novel hand biometric system using invariant descriptors," *IEEE Intl' Conf' on Soft Computing and Pattern Recognition*, pp.261-266, 2014.

[11] S. Sharma, S. R. Dubey, S. K. Singh, R. Saxena, and R. K. Singh, "Identity verification using shape and geometry of human hands," *Expert Systems with Applications.*, vol. 42, pp. 821–842, 2015.

[12] A. El-Sallam, F. Sohel, and M. Bennamoun, "Robust pose invariant shape-based hand recognition," *In Proc. 6th IEEE Conf. Ind. Electron. Appl. (ICIEA)*, Beijing, China, pp. 281–286, 2011.

[13] R.X. Hu, W. Jia, D. Zhang, J. Gui, and L.T. Song, "Hand shape recognition based on coherent distance shape contexts," *Pattern Recognition*, vol. 45, pp. 3348–3359, 2012.

[14] R.M. L.Baena, D. Elizondo, E. López-Rubio, E. J. Palomo, and T.Watson, "Assessment of geometric features for individual identification and verification in biometric hand systems,"*Expert Systems with Applications*, vol. 40, pp. 3580-3594, 2013.

[15] M.H. Wang and Y.K. Chung, "Applications of thermal image and extension theory to biometric personal recognition," *Expert Systems with Applications*, vol. 39, pp. 7132–7137, 2012.

[16] A.K. Jain and N. Duta, "Deformable matching of hand shapes for verification," *In Proc. Intl. Conf. Image Processing*, pp. 857–61, 1999.

[17] B. Zhang, W. Li, P. Qing, and D. Zhang, "Palm-Print Classification by Global Features," *IEEE Trans. on Systems, Man, and Cybernetics: Systems*, vol. 43, no. 2, pp.370-378, March, 2013.

[18] J.M. Guo, Y.F. Liu, M.H. Chu, C.C. Wu, and T.N. Le, "Contact-Free Hand Geometry Identification System," *In Proc. 18th IEEE International Conference on Image Processing*, pp.3185-3188, 2011.

[19] J.M. Guo, C.H.Hsia, Y.F. Liu, J.C.Yu, M.H. Chu, and T.N. Le, "Contact-free hand geometry-based identification system," *Expert Systems with Applications*, vol. 39, pp. 11728-11736, 2012.

[20] A. Morales, M.A. Ferrer, R. Cappelli, D. Maltoni, J. Fierrez, J. Ortega-Garcia, "Synthesis of large scale hand-shape databases for biometric applications," *Pattern Recognition Letters* 68 (1), pp. 183-189, 2015.

[21] D. Santos-Sierra, M.F. Arriaga-Gómez, G. Bailador, C. Sánchez-Ávila, "Low Computational Cost Multilayer Graph-based Segmentation Algorithms for Hand Recognition on Mobile Phones," *Intl. Carnahan Conf. on Security Technology (ICCST)*, pp. 1-5, 2014.

[22] G.V. Lashkia and L. Anthony, "Relevant, Irredundant Feature Selection and Noisy Example Elimination," *IEEE Trans. on SMC: Cybernetics*, vol. 34, no. 2, pp.888-897, April. 2004.

[23] I. Guyon and A. Elisseeff, "An Introduction to Variable and Feature Selection," *Journal of Machine Learning Research,* vol. 3, pp. 1157-1182, 2003.

[24] Z. Sun, G. Bebis, R. Miller, "Object detection using feature subset selection," *Pattern Recognition,* vol. 37, pp. 2165 – 2176, 2004.

[25] H. Peng, F. Long, C. Ding, "Feature Selection Based on Mutual Information: Criteria of Max-Dependency, Max-Relevance, and Min-Redundancy," *IEEE Transactions on Pattern Analysis And Machine Intelligence*, vol. 27, no. 8, pp. 1226 -1238, 2005.

[26] R. Kohavi, G. H. John, "Wrappers for feature subset selection," *Artificial Intelligence,* vol. 97, pp. 273-324, 1997.

[27] P. A. Estévez, M. Tesmer, C. A. Perez, J. M. Zurada, "Normalized Mutual Information Feature Selection," *IEEE Transactions on Neural Networks*, vol. 20, no. 2, pp. 189-201, 2009.

[28] B. Guo and Mark S. Nixon, "Gait Feature Subset Selection by Mutual Information," *IEEE Trans. on Systems, Man, and Cybernetics: Systems and Humans*, vol. 39, no. 1, pp.36-46, Jan., 2009.

[29] T. Zhang, "Adaptive Forward-Backward Greedy Algorithm for Learning Sparse Representations," *IEEE Transactions on Information Theory,* vol. 57, no. 7, pp. 4689 – 4708, 2011.

[30] L. Breiman, "Bagging Predictors," *Machine Learning,* vol. 24, pp. 123-140, 1996.

[31] L. Breiman, "Random forests," *Machine Learning,* vol. 45, no.1, pp. 5-32, 2001.

[32] N. Otsu, "A threshold selection method from gray-level histograms," *IEEE Trans. on Syst. Man, Cybern.,* vol.9 , no.1, pp. 62–66, 1979.

[33] J. Canny,"A computational approach to edge detection," *IEEE Trans. on Pattern Anal. and Mach. Intelligence*, vol. 8, no.6, pp. 679–698, 1986.

[34] A. Kumar, C.M. Wong, H.C. Shen, and A.K. Jain, "Personal authentication using hand images," *Pattern Recognition Letters*, vol. 27, no.13, pp.1478-1486, 2006.

[35] L. Sun, W. Wei, and F. Liu, "A Hand Shape Recognition Method research based on Gaussian Mixture Model," *IEEE Intl. Conf. on Optoelectronics and Image Processing*, pp. 15-19, 2010.

[36] B. P. Nguyen, W.L. Tay, and C.K. Chui, "Robust Biometric Recognition From Palm Depth Images for Gloved Hands," *IEEE Trans. on Human–Machine Systems,* vol. 45, no. 6, pp.799-804, Dec., 2015.